\documentclass[11pt]{article}

\usepackage[margin=1in]{geometry}  
\usepackage{setspace}  
\usepackage{caption}
\usepackage{gensymb}

\usepackage[fleqn]{amsmath}  
\usepackage{amssymb}
\usepackage{empheq} 
\usepackage{bm,upgreek}  

\usepackage{changepage}

\usepackage{url}

\usepackage{algorithmic,algorithm}  

\usepackage{graphicx}  
\usepackage{xspace}

\usepackage{color}  

\usepackage{tikz}
\usetikzlibrary{fit,positioning}
\usetikzlibrary{bayesnet}
\usepackage{wrapfig}

\usepackage{footmisc}
\newcounter{secnum}
\usepackage[all]{nowidow}

\usepackage{titlesec}
\titleformat*{\section}{\Large\bfseries}
\titleformat*{\subsection}{\onehalfspacing\Large}
\titleformat*{\subsubsection}{\large\bfseries}
\titleformat*{\paragraph}{\large\bfseries}
\titleformat*{\subparagraph}{\large\bfseries}

\usepackage{fancyhdr}
\usepackage{indentfirst}
\usepackage{lipsum}  

\usepackage{adjustbox}
\let\oldhash\#%
\DeclareRobustCommand{\#}{\adjustbox{valign=B,totalheight=.35\baselineskip}{\oldhash}}%

\newfont{\namefont}{cmr10 at 12.5pt}

\begin{document}
\vspace*{0cm}
\begin{center}
\noindent{\LARGE Frontiers in Collective Intelligence:\\[.25cm]  A Workshop Report}
\vspace{.5cm}

\rule{0.75\textwidth}{.4pt}
\vspace{.5cm}

\begin{minipage}{.3\textwidth}
\centering
\onehalfspacing
{\namefont Tyler Millhouse}\\ Santa Fe Institute\\ tyler.millhouse@santafe.edu\\
\end{minipage}
\begin{minipage}{.3\textwidth}
\centering
\onehalfspacing
{\namefont Melanie Moses}\\ University of New Mexico\\ melaniem@cs.unm.edu\\
\end{minipage}
\begin{minipage}{.3\textwidth}
\centering
\onehalfspacing
{\namefont Melanie Mitchell}\\ Santa Fe Institute\\ mm@santafe.edu\\
\end{minipage}\vspace{.5cm}

\rule{0.75\textwidth}{.4pt}
\vspace{.5cm}
\begin{quote}
\textbf{Abstract:} In August of 2021, the Santa Fe Institute hosted a workshop on collective intelligence as part of its Foundations of Intelligence project. This project seeks to advance the field of artificial intelligence by promoting interdisciplinary research on the nature of intelligence. The workshop brought together computer scientists, biologists, philosophers, social scientists, and others to share their insights about how intelligence can emerge from interactions among multiple agents---whether those agents be machines, animals, or human beings. In this report, we summarize each of the talks and the subsequent discussions. We also draw out a number of key themes and identify important frontiers for future research.

\end{quote}

\end{center}

\newpage
\tableofcontents
\newpage
\section{Overview}

When building intelligent systems, the need to employ complex systems comprising a large number of more basic components seems inescapable. Brains are composed of billions of neurons, and digital computers are composed of billions of transistors. It is the myriad interactions among these components that give rise to the remarkable larger-scale phenomena of cognition and computation. In this sense, there is no question of whether collective phenomena are relevant to understanding the nature and emergence of intelligence. 
However, to characterize a system as exhibiting collective intelligence is not merely to say that it is composed of many interacting components. It is to say (i) that these components \textit{themselves} possess a degree of agency, intelligence, or ability to react to their environment (including other agents) and (ii) that this fact is central to the emergence of intelligence in the system as a whole. As Levin and Dennett (2020) argue:
\begin{quote}
\onehalfspacing
``There is a profound difference between the ingenious mechanisms designed by human intelligent designers... and the mechanisms designed and assembled by natural selection... Individual cells are not just building blocks, like the basic parts of a ratchet or pump; they have extra competences that turn them into... agents that, thanks to information they have on board, can assist in their own assembly into larger structures.''
\end{quote}
On this view, intelligence is not merely an emergent high-level phenomenon but one that plays a critical role at many levels of organization. To borrow Levin and Dennett's turn of phrase, it's ``cognition all the way down.''

In this workshop, we gathered experts from computer science, biology, social science, and other fields to explain their research and how it bears on larger questions about collective intelligence. For example, which of the mechanisms that enable collective intelligence to emerge (e.g., in social insects and swarm robotics) can inform collective AI? Which frontiers of AI research hold the most promise for advancing large-scale scientific or social efforts to address complex problems (e.g., climate change or epidemic control)? How can we facilitate productive interactions between humans and machines? What kinds of interfaces are best, and what kinds of algorithmic transparency or explainability will most enhance collaborative intelligence between humans and machines? How can we develop collaborative intelligence that promotes important values such as fairness, especially when dealing with collaboration on a large scale? In what follows, we summarize the insights offered by our speakers and attempt to distill the larger themes present in the talks and discussions.

\section{Summaries of Talks and Discussions}

\subsection[``Collective Intelligence in a Computer Model of Analogy-Making'' (Melanie Mitchell)]{``Collective Intelligence in a Computer Model of Analogy-Making''}

\begin{adjustwidth}{1cm}{1cm}
\onehalfspacing
\textit{Melanie Mitchell is the Davis Professor of Complexity at the Santa Fe Institute. Her research focuses on conceptual abstraction, analogy-making, and visual recognition in AI systems.}
\end{adjustwidth}

\noindent Melanie Mitchell presented her ``Copycat'' architecture as a model of collective intelligence. Copycat is a model of abstract perception and analogy making inspired by the kind of collective information processing found in natural systems (e.g., insect colonies, cells, or immune systems). Copycat operates on the domain of \textit{letter string analogies}. Solving these analogies requires one to identify the rule used to transform one string into another and apply that rule to a new string. For example, if one is shown ''abc$\rightarrow$aabbcc'' and asked to complete ``def$\rightarrow$'', one would complete the string with ``ddeeff''. Of course, the rules may also be applied to groups of characters and may include more complicated operations than duplication. For example, ``abcdef$\rightarrow$cbafed'' requires one to realize that the first three and last three letters should be treated as separate sequences and individually reversed. This flexibility with respect to grouping and rules makes solving letter string analogies a surprisingly subtle and challenging task.

To solve these problems, Copycat begins with a network of concepts and employs a number of agents which work in parallel to identify instances of these concepts in the strings. For example, some of the agents might attempt to identify structures (e.g., groups of characters), while other attempt to identify transformations (e.g., reversal) that fit the example given. No single agent settles on a solution, but rather the agents collectively settle on a working solution after a stochastic process of trial, error, and exploration influenced by a global variable ``temperature'' which is determined by the activity of individual agents and which feeds back into that activity by influencing the degree of stochasticity they exhibit. This architecture allows for both bottom up and top down influence; for example, the identification of a plausible group or a plausible rule will bias other agents to find other rules or groups that are consistent with this identification. Hence, the Copycat architecture is an excellent example of collective intelligence emerging from the interactions of many simpler agents.

As Mitchell argues, Copycat incorporates several important ideas relevant to collective intelligence. First, it models high-level cognition as a form of perception, where representations are actively built up over time and where top-down and bottom-up processes interact dynamically. It also continually integrates prior knowledge (e.g., the concept network), and it exhibits an emergent transition from bottom-up stochastic parallel processing to top-down serial deterministic attentive processes as the system settles on a particular coherent structure. This transition also reflects the fact that the system incorporates both sub-symbolic elements (i.e., the stochastic competing agents) with symbolic processing (i.e., the identification and application of symbol manipulation rules). Crucially, Mitchell suggests, these features all depend on the collective nature of the architecture. Moreover, similar architectures (e.g., neuro-symbolic algorithms or probabalistic program synthesis) might help us to address problems in other challenging domains, such as the Abstraction and Reasoning Corpus (or ``ARC'') (Chollet, 2019) and visual situation recognition (Quinn et al., 2018). 

\noindent \textbf{Discussion:}

\noindent The first commenter asked whether the concept network of Copycat could be learned. Mitchell suggested that probabilistic program synthesis might help in this area, but that further work was needed, as this approach is currently limited and requires a great deal of search. The same commenter asked whether temperature worked differently in Copycat as opposed to other algorithms which use a temperature variable to balance exploration and exploitation. Mitchell noted that in Copycat, unlike in other algorithms, temperature is a feedback mechanism.  She  suggested that one could think of Copycat as making a \textit{context-sensitive} exploration/exploitation trade-off using temperature. The next commenter asked what formal letter string analogies had in common with visual situation recognition wherein people must recognize the same situation across many different contexts (e.g., that one can ``walk'' the dog while riding a skateboard and holding the leash). Mitchell said that what matters in both cases is the ability to create transferable representations which re-represent past events in order to apply them to novel contexts. 

\subsection[``Collective Intelligence: Future Directions'' (Jessica Flack)]{``Collective Intelligence: Future Directions''}

\begin{adjustwidth}{1cm}{1cm}
\onehalfspacing
\textit{Jessica Flack is a professor at the Santa Fe Institute and the Director of SFI's Collective Computation Group. Her research focuses on the role of collective computation in the emergence of robust structure and function in both nature and society.}
\end{adjustwidth}

\noindent Jessica Flack outlined the field of collective computation and its relationship to the kinds of problems discussed in collective intelligence. Most broadly, she contends, the study of collective phenomena is the study of pattern formation in many-body systems. This study encompasses many disciplines, ranging from theoretical computer science, to biology, to statistical mechanics. Within this larger area of research, the study of collective computation aims to (i) find common principles underlying diverse examples of collective computation, (ii) provide an organizing framework for relevant sciences, and (iii) develop a general theoretical formalism for characterizing collective computation. 

A key concept in collective computation is the connection between coarser- and finer-grained descriptions, and Flack argues that understanding this connection is central to understanding the computation that underlies collective intelligence. The process of distilling relevant micro-scale information into a smaller set of macro-scale variables is known as ``coarse-graining.'' To take a mundane example, temperature can be understood as the average kinetic energy of the molecules in a solution. Using temperature to represent the solution means that the fine-grained details about the energies of specific molecules are lost, but an important and informative statistic is preserved. In much the same way, organisms and collectives of organisms can help to manage complexity and identify important regularities by distilling the information they receive from their senses into coarser-grained variables of interest. For example, the brain's visual system parses complex and highly detailed visual scenes into macro-scale objects and relationships between them. These higher-level categories are invariant with respect to finer-grained differences (e.g., differences in viewing angle, illumination, etc.) and help us to reliably track salient features of our environment. 

As Flack argues, collective computation has an important role to play in this coarse-graining process. Individual agents have limited samples and limited computational resources. This means that they might not reliably extract the value of a coarse-grained variable. For this reason, collectives can work to aggregate the judgments of individuals into a more accurate and robust estimate of the variable. For example, monkeys have an interest in knowing whether they could successfully use force to resolve a conflict with another monkey. They also have partial/imperfect access to a history of fights between different monkeys, and they can use this history to estimate the relative dominance of different individuals---a coarse-grained representation relative to the full fight history. They also reveal their estimate to others by giving publicly visible subordination signals to monkeys they perceive to be more dominant monkeys. Collectively, we can think of these signalling events as expressing a directed graph of subordination signals that aggregate and communicate each monkey's information about the history of fights. 

The result is a more robust and accurate representation of relative dominance in the group. This representation can be used to avoid needless (and often dangerous) fights between individuals---a serious social problem. In addition, it supports the development of new behaviors that were previously more difficult. For example, if the signals reveal a few individuals that are markedly more dominant, those individuals can more confidently intervene in conflicts between other monkeys since their own dominance is more reliably established. This ``policing'' behavior can benefit a group of monkeys and allows them to solve (at least in part) an important social problem. As Flack argues, and this case illustrates, understanding how groups of individuals collectively compute and aggregate coarse-grained information is central to understanding the kind of collective problem solving of interest to those working in collective intelligence.    

\noindent \textbf{Discussion:}

\noindent The discussion began with a request for clarification about the precise relationship between collective intelligence and collective computation. Flack noted that collective computation is a component of collective intelligence, but not all collective computation constitutes collective intelligence. For example, the framework of collective computation can just as easily investigate the role of collective computation in collective stupidity---or failures of collective intelligence. Another key issue that arose was that of quantifying the degree of decomposability in collective computation. That is, what is the best way to identify and characterize the degree to which a collective computation can be described in terms of the activity of intermediate scale components. Digital computers have a high-degree of modularity and, hence, decomposability, but it is far from clear whether the same is true of computing collectives of animals or neurons.

\subsection[``Collective Intelligence Within the Brain'' (Jeff Hawkins)]{``Collective Intelligence Within the Brain, The Thousand Brains Theory''}

\begin{adjustwidth}{1cm}{1cm}
\onehalfspacing
\textit{Jeff Hawkins is the Co-Founder of Numenta, which seeks to advance artificial intelligence through a better understanding of the neocortex. Previously, he founded and directed the Redwood Neuroscience Institute.}
\end{adjustwidth}

\noindent Jeff Hawkins argued that the brain's cortical columns are a striking example of collective intelligence. These columns of neurons perform similar functions in parallel, resolving disagreements through a voting-like mechanism. The neural representations (i.e., firing patterns) within these columns are sparse---that is, most of the neurons are not active at any given time. The content of these neural representations is encoded by which subset of neurons are firing and how they are firing. This type of neural representation is called a ``sparse distributed representation.'' 

According to Hawkins, these representations are central to the operation of and communication between cortical columns. For example, uncertainty about which of two representations should be active can be represented by activating the \textit{union} of those representations. If it is a dark night, and one cannot tell whether a horse or a cow is standing far away in a field, all the neurons will fire that would have fired for \textit{either} the cow or horse representation. Sparse distributed representations also help to minimize the number of connections needed between columns. If one column would benefit from knowing whether another column thinks there might be a horse within its receptive field, it need not connect to all the neurons that fire when other objects are detected. This would not be the case if the firing patterns were dense---that is, if different neural representations (e.g., ``horse'') involved patterns of firing that included most or all of the relevant neurons. 

Each cortical column is similar to the others in that each attempts to integrate lower-level sensory information into a model of what is in the world. That said, these models are situated within different reference frames for different cortical columns. A reference frame provides a spatial and sensory context within which objects in the world are positioned and characterized. For example, consider two cortical columns that are receiving visual and tactile information (respectively) about a coffee cup. Both might represent the coffee cup as occupying a position in space, but they will differ in the kinds of surface properties (i.e., color or texture) they ascribe to the coffee cup. The same goes for cortical columns which receive different inputs from the same sensory modality (e.g., tactile inputs from the tip of the right index finger and right ring finger). The resulting models will be similar but  situated in reference frames that, as it were, provide an index finger's view or ring finger's view of the world.

These differing perspectives allow the columns to compare their individual models to collectively produce accurate and robust judgments about the environment. As mentioned earlier, cortical columns aggregate their individual models via a voting-like mechanism. The output of these votes, argues Hawkins, is what manifests in consciousness. For example, as one commenter pointed out, this likely gives rise to Gestalt switches in perception like those seen in bi-stable images (e.g., the duck-rabbit). In this way, the cortical columns act as a collective intelligence of ``miniature brains,'' aggregating their judgments about the environment, with different cortical columns making those judgments from within a wide range of reference frames. 



\noindent \textbf{Discussion:}

\noindent The discussion began with some clarifying questions about the voting mechanism described by Hawkins. For example, does the mechanism require a large number of inhibitory connections to quiet the ``losing'' columns and allow a collection of columns to converge on a single model of the world? Hawkins acknowledged the need for some inhibitory connections, but argued that the convergence process is surprisingly quick and that sparsity means that fewer connections are needed to accomplish it than one might imagine. An important point of discussion was how Hawkins's model related to the idea that a (or perhaps \textit{the}) central goal of the brain is to predict future sensory inputs. Hawkins agreed that prediction is central to cortical columns but was not fully convinced of particular interpretations of this fact (e.g., Bayesian interpretations). Other questions highlighted the importance of attention mechanisms in coordinating columns and the general importance of building appropriate reference frames for organizing knowledge.  

\subsection[\hspace{1cm}General Discussion: Day One]{General Discussion:}

Melanie Moses began the general discussion by highlighting some possible areas for further discussion drawn from the day's talks.  These included further clarifying and disambiguating closely related topics like collective, distributed, and collaborative intelligence as well as models, representations, and coarse-grainings. She also noted the importance of balancing top-down and bottom-up influences. Finally, she noted three more specific points: the importance of copying/sharing information, the limitations of individual agents as a motivation for collective intelligence, and the importance of persistent information (e.g., memory) in collective intelligence.

The first attendee to comment asked Jeff Hawkins about how analogy might fit into his model of cortical collective intelligence. Hawkins replied that analogies between models amounted to the kind and degree of overlap between parts of those models. To take a simple example, both staplers and doors have hinges, and familiarity with hinges in either context might be a useful starting point understanding how to interact with the other. The next commenter asked Hawkins and Jessica Flack to comment on Moses's request for clarification on models, representations, and coarse-grainings. Flack noted that coarse-graining is (in her work) a rigorous formal notion that, informally, means a compressed representation of a system that is true to the relevant mechanism. This notion is drawn from physics and connects work in that field to work in cognitive science. Models, in turn, are compressed representations of regularities in systems. Hawkins, in contrast, noted that ``coarse-graining'' is not a common term in his areas of research, but that a model is something quite specific: a neural recreation of the structure of the target system, paradigmatically, a neural representation of an environment or object. How abstract reasoning (e.g., about fair voting laws in democracies) works, he does not know, but he suspects there is a re-appropriation of the basic machinery for more concrete reasoning. 

The next commenter asked Hawkins what he might say about organisms that lack cortical columns. In some cases (e.g., the case of birds), Hawkins argued that neuroscientists have identified analogous structures---that is, structures which seem to carry our similar modeling/learning functions. Hawkins argued that this kind functional similarity is really what matters. To engage and re-engage with objects in their environment, animals must have some means of building a reference frame within which their models of the objects are situated. In other words, its all about discovering/learning the structure of the world and using that structure to guide action. The final commenter concluded with some advice for beginning to resolve some of the semantic ambiguities encountered during the workshop and for learning from other fields. She suggested, in particular, that researchers should listen when experts in other fields say that research in those fields might be relevant to their own. Getting past semantic differences and learning from this research, however, means that researchers must roll up their sleeves and acquire a basic facility with scholarship in other fields. 

\subsection[``How Collective Behavior Unfolds from Individuals'' (Cleotilde Gonzalez)]{``How Collective Behavior Unfolds from Individuals: Interdependence, Aggregation, and Strategy''}

\begin{adjustwidth}{1cm}{1cm}
\onehalfspacing
\textit{Cleotilde (Coty) Gonzalez is a Research Professor at Carnegie Mellon University. Her research focuses on human decision making in dynamic and complex environments.}

\end{adjustwidth}

\noindent Cleotilde Gonzalez discussed her work on how individuals interact in networks given different network structures and different levels of knowledge about others in their network. For example, people in a network might be connected to a single other person, connected in a loop, connected to every other person, etc. Similarly, they might have no knowledge that the consequences of their choices are influenced by the choices of others \textit{or} they might know that these choices matter, know what these choices are, and know how those choices impact them (or anything in between). As Gonzalez argues, what we know about how our choices are entangled with the choices of others shapes what choices we make. Consider, for example, the prisoner's dilemma.\footnote{In the dilemma, two prisoners being interrogated by the police must decide whether to defect and incriminate each other or to cooperate and keep quiet. Both prisoners would get less jail time if they both cooperate than if they both defect. That said, they get the least jail time if they defect while their partner keeps silent and the most jail time if they keep silent while their partner defects. Hence, whether one's partner defects or cooperates, the choice that minimizes jail time is defection. However, this is true for both prisoners and the result (should they adopt this strategy) is that they will both defect and spend more time in jail than if they had both kept quiet. This dynamic changes when we consider repeated interactions where individuals can build trust, punish others for defection, etc.} Gonzalez has found that when individuals know more about their interactions with others, they cooperate at higher rates in the prisoner's dilemma. This matched the predictions made by formal models of reasoning. Naturally, the ability to appreciate the potential benefits of cooperation requires enough information to understand or infer how the prisoner's choices shape their payoffs. 

These results assumed that individuals were paired off when playing the prisoner's dilemma, and it might be natural to think that adopting a more connected network of interactions would not change the basic result. On the contrary, Gonzalez found that the more individuals each person interacted with, the less cooperation there was overall. The explanation, she argues, is that repeated interactions with the same individual reinforce the strategy of cooperation. However, when individuals switch between social partners, they might change their strategy only to switch another social partner whose strategy has not been informed by their past interactions. This limits the ability of individuals to establish patterns of trust and mutual cooperation. Given that cooperation is desirable and given that humans typically interact with many individuals, an important next step for her research (she argues) is to ask what new mechanics must be introduced to reinforce cooperation in the context of complex social networks like our own. 

\noindent\textbf{Discussion:} 

\noindent The discussion participants suggested some interesting ways to extend Gonzalez's work to new areas. For example, might understanding these dynamics help us to develop better artificial collective intelligence or to explain our preference for interacting with individuals in our in-group? Others wondered whether we might understand the network itself as carrying important information. Gonzalez suggested that this characterization might be more apt if individuals learned who it was best to interact with over time. In this way, the connections in the network might encode helpful information (e.g., who should be avoided). Gonzalez identified this as an important area for further study. In addition to learning whom to interact with, another commenter suggested that memory for past interactions with others might allow people to tailor their play to those interactions and (effectively) reproduce the effects of repeated interactions. 
    
\subsection[``No-Bullshit Democracy'' (Henry Farrell)]{``No-Bullshit Democracy''}

\begin{adjustwidth}{1cm}{1cm}
\onehalfspacing
\textit{Henry Farrell is the Stavros Niarchos Foundation Agora Institute Professor of International Affairs at Johns Hopkins University. His work focuses on the relationship between democracy and information, the security consequences of international economic networks, and international political economy
}
\end{adjustwidth}

\noindent Henry Farrell outlined an ambitious research program on the nature of democracy. Farrell argues that current scholarly disputes about democracy fail to do justice to the subject. Pessimists about democracy as a collective problem-solving mechanism emphasize the irrationality of individuals in democracy, while optimists emphasize the fact that an aggregation of many diverse opinions can often outperform expert opinion. The concerns of optimists and pessimists are not necessarily at odds---both could acknowledge that individual opinions are noisy, irrational, or error prone. The question is whether the error-reducing effects of aggregation or the error-inducing effects of individual irrationality tend to dominate. That said, Farrell argues that neither approach captures the variation in democracy---sometimes democracy works well and sometimes it works poorly. Research should attempt to understand what factors contribute to the successes \textit{and} failures of democracy rather than arguing for or against democracy \textit{tout court}.

To gain this understanding, Farrell argues that we should investigate democracy as a form of collective intelligence which aims to solve difficult social problems (e.g., navigating disagreements about core values in crafting effective public policy). Farrell identifies three areas of research central to this project, which he is currently exploring with collaborators. First, there is the issue of how democratic problem solving works on small scales and how group structure contributes to (or detracts from) collective problem solving. Of particular importance here are questions about diversity within groups, including both the extent of common ground between members as well as the comfort of members with dissent and disagreement. Once the contribution of such factors is understood at a small scale, researchers should investigate how these lessons scale with group size.  

The next two research areas Farrell proposes take a macro-scale view of democracy. First, there is the issue of how democratic institutions evolve over time. Of particular interest here are the opportunities for individuals to access and participate in these institutions and how this participation might shape the course of their evolution. Second, there is the issue of which factors contribute to the stability or instability of democratic and non-democratic governments. These factors might include everything from access to information, belief that elections are fair and competitive, norms supporting the peaceful transfer of power, etc. 

Farrell also proposes two hypotheses about what research in these areas might uncover. For example, it seems likely that the democratic and non-democratic governments will fare differently given open access to information on the part of citizens---with the former being stabilized by greater access to information and the latter being destabilized. Farrell also proposes that democratic institutions whose participants exhibit greater ideological diversity and have greater access to participating in those institutions will tend to adopt new policies more readily as diversity fosters innovation and access enables change. 

\noindent\textbf{Discussion:} 

\noindent The discussion began with the question of how to define democracy. Is voting on politicians or policies enough? Farrell offered a rough and ready definition of democracy that is more substantive. Democracy is a system which is open to a variety of perspectives and is able to harness information from those perspectives in order to solve social problems. This definition eschews the details of voting, representative government, etc.\ and focuses on the core idea of integrating information from many perspectives to enhance decision-making. Another commenter suggested thinking about democracy in terms of the norms that communities possess which enable fruitful collective problem-solving. Farrell supported this idea and regarded it as complementary to his own. Another commenter wondered where a connection to truth and reality fit into democracy, given its emphasis on integrated diverse views some of which are further from reality than others (especially with respect to individuals' acceptance of well-supported science). Farrell suggested that it would be hard, say, to return to a norm of deference to science, but that it might be better to communicate the messiness and uncertainty of science in a way that doesn't make science seem brittle and untrustworthy as scientists revise and refine their conclusions. Finally, one commenter worried that democratic institutions might foster dissent-averse conformity, while another worried that democracy might not be able to reconcile or integrate highly disparate values.

\subsection[``Efficient organization without too much overhead'' (Anna Dornhaus)]{``Efficient organization without too much overhead: When to reduce communication for better group flexibility''}

\begin{adjustwidth}{1cm}{1cm}
\onehalfspacing
\textit{Anna Dornhaus is a Professor at the University of Arizona, where she directs the Social Insect Lab. Her research focuses on collective problem-solving strategies as well as efficiency, flexibility, and robustness in complex systems.}
\end{adjustwidth}

\noindent Anna Dornhaus summarized important lessons on communication in complex adaptive systems drawn from her research on insect foraging behavior. In the context of this research, Dornhaus understands a complex adaptive system to be a system in which the behaviors of individual components have been selected for some collective outcome. For example, the behavior of cells in animal bodies has been modified by natural selection for the reproductive success of the organism. In contrast, the behavior of economic agents might give rise to collective outcomes---even desirable ones---despite the fact that the behavior of individual agents is selected for non-collective outcomes (e.g., the maximization of their own utility function).\footnote{For clarity, this evolutionary biological account of ``complex adaptive systems'' contrasts with how the term is applied in other areas of complex systems science. In the latter context, ``adaptive'' refers to the ability of a system to adapt to changing circumstances. This includes, but is not limited to, adaptation over evolutionary time due to selection. It also includes the ways that a system (e.g., a social group of humans) responds to changes within the lifetime of its members (e.g., moving to a more fertile location for farming) or within the lifetime of the system (e.g., changing social norms or methods of agriculture)(Gell-Mann, 1994). This is not to critique either account, but rather to highlight an important difference in emphasis.} 

Turning to the issue of communication, non-deceptive communication might seem like a form of cooperation that would benefit collectives, but not all communication is actually selected for collective outcomes. For example, a peacock might communicate an honest signal about its fitness by having a showy tail, and this information might benefit potential mates. Nevertheless, this behavior was not selected for any collective outcome, but rather because it benefits the sender of the signal. In contrast, the more modest waggle dance of a bee serves to convey important information about the location of food sources to other members of the colony, and this behavior appears to have been selected for the collective reproductive success of the hive. In general, Dornhaus argues, non-cooperative signalling tends to be loud and showy, while cooperative communication tends to be less showy but more information-rich. 

Even cooperative communication, however, may not be desirable in every context, and the contexts in which it is beneficial give us clues about the function of communication within a given system. For example, Dornhaus notes that larger bee colonies seem to benefit more from communication than smaller colonies. The reason for this seems to be related to the costs of communication. Each bee that forages faces a choice to go out and look for food or wait for another bee to scout out food and report back. If food is easy to find, then it might be worth finding it yourself rather than waiting for another bee to go out, find the food, fly back, and report. This waiting period will be less, however, when there are many bees to serve as scouts, since it is more likely that at least one will stumble upon food very quickly and report back. 

Other insects tune their sensitivity to signals from others rather than merely waiting or not waiting for those signals to arrive. For example, different species of ants follow different patterns in foraging for food: in some species, ants spread across multiple food sources while in others, many ants converge on a single food source. The latter species tend to be more physically dominant, and by arriving in force, they can more easily drive off other ants. Less dominant ants, naturally, benefit from having ready alternatives in case they are driven off. This behavior is modulated by how ants respond to pheromone trails left by other ants. If this response is such that a slightly stronger trail is much more likely to be followed, then ants primarily follow this trail, adding their own pheromones as they do, thereby reinforcing the original signal. These ants quickly converge on a single food source. In contrast, if ants are only somewhat more likely to follow a slightly stronger signal, then they tend to spread out to multiple food sources. Ultimately, the exact forms of communication that emerge in complex adaptive systems can tell us a great deal about the nature of those systems (e.g., their population and foraging strategies) and how they are situated in particular contexts and environments (e.g., the presence of physically dominant competitors).  
        
\noindent\textbf{Discussion:} 

\noindent The discussion focused on another set of trade-offs mentioned in the talk---those involving the proportion of specialists and generalists in the colony and the extent of the division of labor in the colony. For example, in some of Dornhaus' work, the costs of switching between tasks (e.g., caring for larvae versus scouting) favored the division of labor. That said, Dornhaus emphasized a broader set of factors, any one of which \textit{could} give rise to specialization and the division of labor.\footnote{Here we distinguish between specialization and division of labor, since specialization in insect colonies might involve significant differences in morphology between specialists and generalists, rather than merely the specialization of individuals on particular tasks (as in human economies).} The difficulty, she argued, was in assessing these factors in a real-world system such that one could identify the reasons for a particular level of specialization or division. For example, how might we measure the difficulty of reallocating generalists from one task to another? It is not clear. What is clearer is that there is a close connection between the size, structure, physiology, and environment of an organism and the trade-offs it makes with respect to communication, specialization, and division of labor. Understanding these connections is an important area for future study.  

\subsection[\hspace{1cm}General Discussion: Day Two]{General Discussion:}

Melanie Mitchell began the general discussion by highlighting some important open questions raised by the talks. For example, all the talks touched on the information possessed and shared by individuals within groups and how this information facilitated problem solving. However, in some talks (e.g., Cleotilde Gonzalez's) emphasized the fact that the resulting solutions are realized in the knowledge and behavior of many individuals. Mitchell asked how much of what groups know or learn is realized by, say, the pattern of connections between individuals or other aspects of group structure. Further, how do specific variations in group structure affect learning, problem solving, and cooperation? Access to information about others, degree of hierarchical organization, division of labor, and other factors might all play critical roles in the emergent behavior of the group. 

One issue raised by Anna Dornhaus was the idea that complex adaptive systems, understood from the standpoint of an evolutionary biologist, are those systems whose individuals members have been adapted in some way for a collective outcome. It is controversial whether the course of human evolution has been shaped by selection for group-level outcomes, but Mitchell asked whether we might find that certain cognitive biases which seem detrimental to individuals (e.g., confirmation bias) might be better explained if we consider how they facilitate cooperation or other desirable collective outcomes. Turning to ideas raised by Henry Farrell, Mitchell suggested further discussion about the factors that help to stabilize democracies and whether there might be analogs of these factors in other collective intelligences. She also asked whether stabilizing influences of this sort can be expected to scale from small human groups to large-scale human societies. 

The broader discussion opened with a question about the centrality of consensus to collective intelligence. In the context of, say, democratic decision-making, there may often be times that call for some form of consensus or, at least, some way of settling on a policy despite disagreement (e.g., enacting either this or that tax policy but not both). However, it is less clear that we should want, say, economic decision-making to have strong consensus-forming mechanisms. An economy might be more robust if investors diversify and businesses explore a range of new products and services. For this reason, the commenter suggested that we also consider how to optimize collectives for intelligent behavior that does not prioritize consensus. 

The next commenter asked Farrell why politics is so different from science and engineering. Problems in these areas are not solved by writing op-eds, garnering public support, or voting. Given the rapid progress of these fields, what does this tell us about the comparative intelligence of collective political decision-making? Farrell responded that the problems are importantly different in politics. For example, values play a much larger role, are harder to formalize than the aims of scientific research, and are much more difficult to reconcile when in conflict. Another commenter indirectly echoed these thoughts, noting that even with agents who are pre-disposed toward cooperation, there are hard design questions about how to channel cooperation in helpful ways (e.g., by tuning response to pheromones). Another commenter suggested that the goals of individual humans are shaped by biology, which was largely unconcerned by what modern people regard as desirable social outcomes. This means that our individual goals might often be in tension with the social outcomes we (abstractly) endorse. 

The next commenter spoke to the issue of where solutions are represented in collectives, and argued that the choice between the ``in many individuals'' and ``in the connections between individuals'' views might be a false dichotomy. For example, collective outcomes are often the product of human action but not human design (Ferguson, 1782). This means that the goals behind individual human actions (e.g., making or saving money) may be orthogonal to the collective outcomes that result (e.g., the reduced consumption of scarce goods without the need for rationing due to rising prices). This allows us to take a collective level view, evaluating the individual goals and actions as selected for (or not selected for) a  particular collective outcome. That said, it also allows us to take an individual level view, evaluating how well optimized each individual's actions are for achieving their goals. Both are valid and important perspectives to take. Nevertheless, another commenter worried that whether actions or goals are optimal for bringing about some collective effect is a different question from whether those actions or goals have been subject to a process of selection. The latter might be importantly different in practice. For example, the hypothesis that certain ant behaviors have been selected for a particular collective outcome might predict the discovery of other traits selected for that outcome. The claim that ant behaviors happen to give rise to certain collective outcomes \textit{as if} those behaviors were selected for that outcome would not make similar predictions.

The discussion next turned to the idea, proposed by a commenter, that many mechanisms of collective intelligence might involve shifting group boundaries (e.g., the formation of political or neural coalitions). Another commenter noted that these kinds of coalitions are (in human societies) often goal-oriented (e.g., facing an external threat). Others, however, argued that this could not fully explain cooperation in these cases since the mutual perception of a threat already presupposes a mechanism of consensus formation that may or may not be present. After all, the mere presence of an external threat does not guarantee that individuals in a society will be able to identify and collectively agree upon its existence. For example, global warming is a threat to human society, but many efforts at unified action against it have failed because individuals disagree about the existence and severity of the threat. Also, the benefits of a shared goal can be reduced by the fact that some will benefit more from accomplishing that goal than others and some will need to make greater sacrifices than others to accomplish the goal. Again, different individuals, regions, and industries have more or less to gain (and lose) from combating global warming. 

\subsection[``When (and How) Crowds Can be Less Than Wise'' (James Marshall)]{``When (and How) Crowds Can be Less Than Wise''}

\begin{adjustwidth}{1cm}{1cm}
\onehalfspacing
\textit{James Marshall is the CSO and Co-Founder of Opteran Technologies and a Professor at the University of Sheffield. His research focuses on biologically-inspired algorithms, decision theory, social evolution, and social insect behavior.}
\end{adjustwidth}

James Marshall explored the ways natural and artificial populations of information-accumulating agents solve decision problems under differing conditions and capabilities. Two salient points from his talk are (i) that the misapplication of decision theory can lead to failed predictions or bad decisions, and (ii) that the simple, elegant decision-making strategies of neurons, honey bees, and collections of robots share important similarities and can enable effective collective decisions in a variety of real-world problems.

The talk started with an illustration of how collective decisions can be made by a population of neurons in a mammalian brain. A monkey or a human is shown a field of moving dots on a screen. Some of the dots on the screen are moving in the same direction while the rest are moving randomly, such that the field displays coherent but noisy motion. The population of neurons must then decide if the majority of dots are generally moving left or right. The statistically optimal solution to this decision problem can be modeled simply and demonstrates a speed-accuracy trade-off based on Brownian motion toward each of the two possible events, with a decision made based on the log likelihood of each event.  An important component of the neural behavior in this system is mutual inhibition; neurons from one population (e.g., those voting for dots moving left) will proportionally inhibit neurons from the other population (voting for dots moving right). 

This balanced mutual inhibition model also explains other collective decision-making processes, particularly colonies of ants and bees that accumulate evidence and then make a collective decision that all agents then follow. In a manner similar to neurons, honey bees make decisions as competing populations of evidence-accumulators. The theoretical framework suggested that in order to optimally trade off speed and accuracy in choosing a hive site, bees engaged in a waggle dance would require a method of mutual inhibition. That ``stop signal'' was discovered empirically.

The talk then showed some shortcomings of ``the wisdom of the crowd''. Condorcet’s Jury Theorem describes binary decisions in populations. The theorem states that under the assumption that the opinions or individuals are independent and have equal positive decision accuracy, groups make better decisions than individuals if individual accuracy is above one-half. Moreover, as group size increases, the accuracy of the decision approaches 1. Marshall shows that---in many if not most real-world situations---these predictions are based on a false assumption that errors among agents are independent. Specifically, it fails to account for the tradeoff between false positives and false negatives. He used as an example meerkats which collectively forage with sentinels that issue an alarm call when predators are present. False positives are less harmful than false negatives because failure to detect a predator can be deadly while a false negative costs only a brief pause in foraging.  

More generally, majority votes typically fail in two ways: first, when the cost of false positives and false negatives are not equal, the majority vote does not converge to the optimal answer. Additionally, in some cases, the accuracy of the group is lower than that of the average individual. Marshall showed that setting a quorum threshold (rather than simple majority vote) escapes the problems introduced by the false positive/false negative trade-off curve so that the collective is at least as good as the average individual. Marshall and colleagues extended this finding to explicitly state the conditions under which quorums will perform better than majority voting and when majority voting will perform worse than individuals.

In Marshall’s final example, he considered the effect of an unequal distribution of information among agents. Many natural systems, such as neurons and honey bees, make decisions with agents that have incomplete and differing access to accurate information. Marshall demonstrated decision-making capabilities of swarms of simple mobile robots. Small populations of randomly-moving and information-sharing Kilobots were able to collectively discover the ground truth about which home site in their arena was best when robots were only allowed to communicate locally with other robots they came into  direct contact with. Additionally, they could accurately track changes in the ground truth. However, the same robots given the same decision problems, but allowed to communicate information globally, were much less likely to reach an accurate decision. This is fundamentally due to increasingly asymmetric interaction rates between different types of agents as population size increases. 

\noindent\textbf{Discussion:} 

The first commenter asked how sensitive optimal policies (e.g., optimal quorum thresholds) were to changes in context. Marshall was not sure, but argued that the kinds of policies chosen by evolution would, in addition other considerations, be optimized for robustness. The next commenter asked Marshall to say more about bees' use of stop signals and the competitions among individuals that result in collective decisions. Marshall noted that his models had predicted that bees would use some kind of behavior to encourage others to change their minds, but that the dynamics of this process were not as he initially expected. Marshall looked at cases where decisions were between equally good alternatives, hoping to see more protracted conflicts and more extensive use of the stop signal. However, in such cases, it is actually best to quickly identify the fact that the options are equally good and choose one at random. This led to refined models of how bees arrives at a consensus. The commenter followed up by asking what factors are relevant to predicting the optimal level of communication for a collective. Marshall noted that this was fairly easy in simple cases, but that matters were complicated by the fact that performance can fall off quite rapidly with relatively small increases in communication. The final commenter asked whether the kinds of dynamics he studies would change when collectives comprise more sophisticated agents. Marshall noted that pairs of humans can make better decisions when they communicate both their preferred decision and their confidence levels about it, suggesting that sophisticated agents could better leverage more complicated modes of communication. 

\subsection[``Beyond Machines of Flesh and Blood'' (Jacob Foster)]{``Beyond Machines of Flesh and Blood: Toward New Paradigms for Human-Compatible Collective Intelligence''}

\begin{adjustwidth}{1cm}{1cm}
\onehalfspacing
\textit{Jacob Foster is an Associate Professor at the University of California Los Angeles. His research focuses on social theory, computational social science, networks, complex systems, and cognition and culture.}
\end{adjustwidth}

\noindent Jacob Foster outlined an ambitious proposal about how to re-frame the foundations of social theory in terms of collective intelligence. Foster argues that for many of the same reasons that AI engineers are seeking to create ``human compatible'' artificial intelligence (Russell, 2019), we should approach social theory as an exercise in creating human compatible collective intelligence. First, however, it is important to understand collective intelligence (and intelligence in general) before applying these lessons to the particular kind of collective intelligence present in human societies. To this end, Foster and collaborators are seeking design principles that recur in collective intelligence, and have begun to outline several of these principles.

The first is that all intelligence is an emergent property of complex adaptive systems. Further, intelligent systems comprise diverse subsystems which (i) possess some degree of agency and intelligence and (ii) are composed (and re-composed) to meet the demands of particular tasks. The activity of these subsystems is coordinated by mechanisms which guide their collective behavior toward adaptive outcomes. These mechanisms are engaged in a kind of ``reverse game theory'' whereby they attempt (either in evolutionary or learning time) to construct circumstances wherein the interaction of subsystems leads to maximally adaptive outcomes. The overall picture of intelligent systems, then, is one in which complex adaptive systems dynamically reallocate subsystems to meet the demands of their circumstances. This means that particular configurations of subsystems may be quite ephemeral. For example, the brain might recruit assemblies of neurons in order to solve a particular problem, and never recruit the same assemblies in the exact same way again. Intelligence, on this view, is a kind of process rather than a static capacity. 

As Foster argues, this suggests several plausible conjectures. For example, we should expect intelligent systems to have evolved in ways that make their subsystems composable. However, we should also expect this compositionality of subsystems to have co-evolved with mechanism design, leading to important trade-offs in the features of subsystems. For example, there may be an important trade-off between the price of anarchy and the repurposability of subsystems. The ``price of anarchy'' refers to the reduction in performance from not coordinating the activity of individual agents.\footnote{This concept originated in game theory, where it refers to difference between a worst case equilibrium and the most socially desirable outcome (Koutsoupias \& Papadimitriou, 2009).} For example, for every additional traffic sign that is posted, there will be a benefit associated with improved coordination of traffic (i.e., a reduction in the price of anarchy). The trade-off here is with the cost of posting/maintaining traffic signs. Similarly, suppose an intelligent system's subsystems have specialized via learning. The more competent those subsystems are the less costly it will be to coordinate their behavior within their domain of specialization. While this makes it easier to reduce anarchy and its associated costs, it might also make the system less flexible in adapting to novel circumstances. Hence, it may be worth paying (at least some of) the price of anarchy in order to secure this kind of flexibility. 

Hence, there are conflicting reasons to shape the individual subsystems to fit a particular task and to promote diversity, generality, and flexibility. As Foster argues, we should think carefully about this trade-off when designing social institutions. For ethical reasons, we should be hesitant to create institutions that re-shape individuals to achieve desirable outcomes, and we should be wary of how existing social institutions (e.g., governments, social media, etc.) are already doing so. Instead, we should find ways of enhancing the capacity, diversity, and autonomy of individuals and engineer mechanisms for creating circumstances under which the interactions of these individuals result in desirable outcomes. Human compatible collective intelligence, then, is collective intelligence that harnesses the autonomy and diversity of human beings to achieve desirable outcomes.          
        
\noindent\textbf{Discussion:} 

\noindent The discussion began by touching on two areas of Foster's presentation---the idea that intelligent systems are hierarchically composed of intelligent subsystems and that intelligent systems exhibit ``cognition all the way down.'' The commenter suggested that this form of ``structure at all levels'' might be a middle ground between the kind of top-down, bottom-up dichotomy one might see in Foster's presentation of the relevant policy alternatives. Foster largely agreed with the importance of this observation. Another commenter asked about what kinds of cases might demand a more top-down approach. Foster replied that it would depend on the trade-offs involved in each case, but that in general he felt that too many of the costs of top-down approaches were hidden. For example, suppose a social media company develops and implements a transparent and interpretable content recommendation algorithm. These selling points would not guarantee that the values judgments made in the design and application of this algorithm (however transparent or interpretable) are actually good for the network or its members---much less when applied in a one-size-fits all methods to diverse individuals. Finally, others asked about the concept of human compatibility and whether Foster's recommendations of bottom-up organization were needed correctives or an end in themselves. Foster argued that his primary concern is with developing a richer picture of human compatibility that reaches beyond human preferences to considerations of human capacity and human flourishing. This perspective does not require that there be no top-down structure, but it does requires us to count any limitations it places on individual diversity and autonomy as a significant cost.

\subsection[``Applied Collective Intelligence to Combat Online Misinformation'' (Joe Bak-Coleman)]{``Applied Collective Intelligence to Combat Online Misinformation''}

\begin{adjustwidth}{1cm}{1cm}
\onehalfspacing
\textit{Joe Bak-Coleman is a Postdoctoral Fellow at the University of Washington. His research focuses on how individual behavior gives rise to collective action and how this process is affected by communication technology.}
\end{adjustwidth}

Joe Bak-Coleman surveyed some important factors contributing to online misinformation and the strategies we might use to combat them. While complex systems are often thought of as adaptive and resilient to perturbations, this resilience is finite. For systems pushed beyond their breaking point, failures can be catastrophic. For example, during the record cold and snow in early 2021, the Texas power grid collapsed under the unprecedented strain. The same holds for systems that exhibit collective intelligence. For example, if the pheromone trails left by ants form a loop, the ants can follow that loop in circles endlessly, leading in some cases to starvation. As Bak-Coleman argues, the success of these complex systems is context dependent and large changes in context can lead to failure. 

Human communication evolved in the context of small hunter-gatherer groups to solve local problems using vocalization and gesture. As communication technologies improved, human beings could communicate with greater fidelity, but the ability to use mass communication to transmit one's ideas to a wide audience (e.g., via broadcast television) remained inaccessible to most humans---blocked by costs and gatekeeping mechanisms. This was still a radical departure from our environment of evolutionary adaptedness, but the full effect of the changes was blunted by such barriers. The internet allowed even greater fidelity of transmission and even larger audiences, but it also rapidly dismantled the gatekeeping mechanisms that reduced the viral spread of misinformation. Worse, these changes were accompanied by algorithms that tended to promote content in ways that unintentionally encouraged the spread of misinformation, reinforced extremism, etc.

Given the consequences of misinformation and the questionable ability of human social networks to adapt to this shock, Bak-Coleman argues that we should seriously consider implementing some set of interventions to mitigate the spread of misinformation. He considers several strategies for doing so: removing misleading content, slowing misleading content, banning the biggest spreaders of misleading content, and adding ``nudges'' to misleading content that warn users and direct them to authoritative information. Unfortunately, a clear picture of whether and how well these methods work is still emerging. As Bak-Coleman describes, all of these methods (if pursued individually) would likely require draconian levels of enforcement or unrealistic level of effectiveness in order to work as desired. Fortunately, a multi-strategy approach which combines a moderate amount of each strategy looks considerably more promising. 

\noindent\textbf{Discussion:} 

The discussion began with a question about whether the attitudes, motives, etc. that drive the spread of misinformation might be harnessed by the right sort of institutions. For example, in a context where private property is protected, the drive to acquire wealth can be channelled into productive activity that (however imperfectly) benefits society at large. Bak-Coleman agreed that getting the right incentives (e.g., reducing the profitability of spreading misinformation) and the right network structure (e.g., encouraging local interactions where offline relationships hold people accountable) could substantially help matters. Another commenter worried that the problem of misinformation was, perhaps, more a problem of super-spreaders with huge audiences than ordinary people on social networks. On this view, the problem is less about a truly collective effect and really a problem of a few bad actors within a social network. Bak-Coleman agreed to some extent, but noted that the algorithms are designed to promote ``engagement'' and that this is assessed at the level of individuals. Hence, the emergence of highly influential accounts is not unrelated to the activity of individuals. Another commenter wondered whether the discussion was focusing too much on technical problems to the exclusion of deeper political and social ones (e.g., voter suppression). Bak-Coleman acknowledged the importance of these issues, but also noted that these are areas where real design decisions are being made by social media companies, and there is considerable value in ensuring that they make the right decisions.     

\subsection[\hspace{1cm}General Discussion: Day Three]{General Discussion:}

Tyler Millhouse began by outlining a few themes of the workshop as a whole and of the day's talks in particular. He first noted several different concepts that had been mentioned during the workshop, including collective information sharing and aggregation, collective computing, collective decision making, collective problem solving, etc. These all seem relevant to collective intelligence---as perspectives on what intelligent collectives are doing and as subjects of study in their own right. Millhouse asked to what extent these could be individually investigated and how participants thought they might need to be integrated into a general account of collective intelligence. Millhouse next asked when collectives should be counted as intelligent and whether ``intelligence'' in this sense differed from the sense in which it is attributed to individuals. These questions, of course, are complicated by the possibility of forms of collective intelligence that exist within individual organisms (e.g., between cortical columns). Finally, Millhouse asked about the connections between collective intelligence as discussed in the workshop and the kind of temporally-extended collective intelligence exhibited by evolving populations (discussed in the previous workshop). 

Millhouse, next advanced particular points of discussion for each of the talks. For James Marshall's talk, it seemed that a key point was how finer or coarser grained analyses of success could shape what we take to be the optimal organization for a group. For Jacob Foster's talk, the highlight was a substantive notion of intelligence applicable to collectives and the introduction of richer notions of success in navigating AI risk. Finally, for Joseph Bak-Coleman's talk, the idea of mechanisms of accountability and the value of keeping interactions local stood out as possible starting points for discussion. 

The first commenter noted that in human society (at least), overlapping collectives can inter-regulate. For example, if people are members of different political parties but attend the same church, there will be increased social costs to disagreeing too acrimoniously about politics. The commenter noted that for online social media, many of the regulatory difficulties we observe might arise because divisions on these platforms are aligning with divisions offline, resulting in little common ground and, hence, little shared interest in finding ways to constrain the worst excesses of online disagreements.

The next commenter added some nuance to the notion of misinformation by suggesting that we should be more concerned about whether public opinion is well-calibrated or not. There will always be people whose guesses about the weight of the steer at the county fair are wildly off, but this is consistent with the median guess being quite reliable. This is importantly different from a system in which people \textit{in aggregate} are consistently off the mark. For example, if online interactions are driving political polarization, most individuals may be driven to extreme positions rather than converging to some better calibrated moderate position. 

Another commenter noted that much of the recent distrust of experts and expertise may result from the fact that scientific findings and the demands of policy-making interact in unproductive ways. For example, it might occasionally be important to rapidly address some emergent situation (e.g., a pandemic). Not all the measures will be effective or (in the final analysis) advisable, but the need to implement our best guess about the right response could lead scientists to oversell their certainty in the effectiveness of any particular intervention. Another commenter added that certainty is also a powerful rhetorical technique over and above any particular emergency. In any case, the primary worry was that when some of these confident recommendations are reversed, trust in science is eroded. 

That said, it remains unclear whether better presenting the reality of scientific uncertainty would fully resolve the difficult decisions we have to make as a society. Even where the science is in fact clear, there are deep disagreements about core values. Each individual has, during the COVID pandemic, struck some balance between the risks of certain behaviors to themselves and others and the benefits they will have to forego to avoid these risks. Most would agree that shutting down the economy to save a single individual \textit{would not} be worth the cost, and few would disagree that shutting down the economy to save all but one individual \textit{would} be worth the costs. Somewhere between these extremes, real disagreements about what is most valuable will come into play. Nevertheless, there may be better or worse ways of navigating these disagreements, and (as one commenter suggested) many strategies for productive disagreement have been proposed and tried by members of the ``rationality movement'' which seeks to shape individual reasoning and intellectual discourse in order to avoid cognitive biases and fallacious reasoning. 

\section{Conclusion}

While the workshop brought together researchers from a diverse range of fields, several key themes and open questions emerged from the talks and discussions. Overall, the workshop reinforced the idea that understanding collective intelligence is a key part of understanding intelligence in general. The speakers not only showed that collective intelligence is a plausible approach to solving problems in AI and human communities, but also that collective problem-solving is a common and highly successful feature of biological systems. That said, the interdisciplinary nature of the workshop also revealed the diversity of perspectives at play in the study of collective intelligence, ranging from differences in terminology to differences in subject matter and research methods. Incorporating these diverse perspectives into a more unified framework will both advance our understanding of collective intelligence and facilitate interdisciplinary research on the subject. 

\subsection{Optimization and Collectivity}

One of the principal themes of the workshop was that collective phenomena can arise in systems that  exhibit collectivity at different levels of organization and whose components exhibit different degrees of optimization for those phenomena. For example, the talks illustrated that phenomenon of collective intelligence can emerge from the interactions of many individual agents (as in an insect colony) and from the interactions of components within a larger cognitive system (as in the cortical columns of mammalian brains). The organization of these components and the components themselves can also exhibit different degrees of complexity. In many contexts, emergent phenomena in complex systems are characterized as emerging from the interactions of many simple components. While this kind of emergence is striking, Jessica Flack  emphasized  the complexity of individuals in collectively intelligent systems. Humans, animals, and even individual cells are quite complex, and they are often organized in non-trivial ways, with network structure playing an important roles (as emphasized by Coty Gonzalez).  

Further, the components contributing to collective intelligence and other collective phenomena can be optimized to different degrees for the production of those phenomena. For example, there are a number of interesting phenomena that emerge from interactions between humans and interactions between ants, but it seems clear that the individual behaviors of ants are more finely-tuned by natural selection for the production of those phenomena than are the behaviors of humans. This proved to be a point of disagreement in the methodology of different disciplines, with insect biologist Anna Dornhaus emphasizing the fact of natural selection acting on groups in the case of social insect colonies. In contrast, some working in the social sciences were more interested in how (and how well) individual behaviors produce some collective outcome than in whether those behaviors were selected for those outcomes. Of course, as both Dornhaus and Jacob Foster noted, selection for specific individual behaviors does not necessarily imply specialization, and the balance of generalists and specialists is itself something that can be evolved or designed to meet particular goals. Considering all these perspectives, it seems fair to say that the question of whether and how the components of a collective intelligence have been shaped by selection is a substantive one that has important implications for the analysis and function of collective intelligence. For this reason, the field would likely benefit from a more systematic framework for understanding the role of selection in designing collective intelligences. 


\subsection{The Benefits (and Costs) of Communication}

Another major theme was the importance of tuning communication between the components of collective intelligences. For example, in the context of insect colonies, Anna Dornhaus explained the importance of communication in spreading information about food sources among foragers. The value of this information, however, must be balanced against the costs to waiting at the colony for information rather than going out searching for food. Similarly, James Marshall noted cases in which  local communication is beneficial, but global communication is detrimental to foraging behavior. In a much different context, Joe Bak-Coleman noted that misinformation on Twitter is disproportionately spread by a relatively small group of serial offenders with large followings, suggesting that one vector for improving the quality of information on the internet would be to target the relatively small number of worst offenders. Hence, while important, communication must be finely tuned, both with respect to the amount of communication that occurs and the structure of communicative interactions. 

\subsection{Institutions, Norms, and Mechanisms}

Another key idea from the workshop was the importance of mechanisms, broadly construed, which coordinate the activity of the components of collective intelligence. As Jacob Foster described, mechanism design amounts to a kind of reverse game theory where the conditions of interactions are engineered to produce desirable outcomes. In the context of cognitive systems, these mechanisms are responsible (among other things) for the strategic re-allocation of cognitive resources. In the context of insect colonies, these mechanisms involve different modes of communication and different processes for dividing labor among colony-members. In the context of the wider discussion about political systems and social networks, these mechanisms involve institutions, rules, norms, and shared beliefs. In each case, modifying these mechanisms can make a big difference to the collective outcomes produced. 

\subsection{Lessons for Artificial Intelligence}

Each of the workshop's major themes suggests an avenue of research for scientists developing collective artificial intelligence. The first suggests approaching the design of individuals, components, and subsystems as a kind of optimization problem---perhaps one to which evolutionary algorithms or other methods might be applied. The second recommends careful attention to channels of communication and cautions against the idea that unrestricted communication is desirable. The last suggests greater attention to the factors that structure interactions, and it recommends closer attention to the normative judgments that guide mechanism design. Taken together, these themes highlight the expansive possibilities and promising frontiers for future research on collective intelligence.

\newpage
\section*{Acknowledgments}
The workshop was funded by a grant from the National Science Foundation (\#2020103) as part of the Foundations of Intelligence in Natural and Artificial Systems project at the Santa Fe Institute.

\section{References}

\begin{flushleft}

\begin{list}{}
{\leftmargin=1em \itemindent=-1em \itemsep=-.4em}

\item Ferguson, A. (1782). \textit{An essay on the history of civil society}. London: Caddell, Creech, \& Bell. 

\item Gell-Mann, M. (1994). Complex adaptive systems. In H. Morowitz and J.L. Singer (eds.) \textit{The mind, the brain, and complex adaptive systems}. New York: Routlegde.

\item Koutsoupias, E. and Papadimitriou, C. (2009). Worst-case equilibria. \textit{Computer science review}, 3(2). 65-69.

\item Levin, M. and Dennett, D. (2020). Cognition all the way down. \textit{Aeon}. \url{https://aeon.co/essays/how-to-understand-cells-tissues-and-organisms-as-agents-with-agendas}.

\item Quinn, M. H., Conser, E., Witte, J. M., and Mitchell, M. (2018). Semantic image retrieval via active grounding of visual situations . In \textit{Proceedings of the 12th International Conference on Semantic Computing}. IEEE. 

\item Russell, S. (2019). \textit{Human compatible}. New York: Viking.

\end{list}
\end{flushleft}

\end{document}